%% file: main.tex
\documentclass[10pt,twocolumn,letterpaper]{article}

\usepackage{iccv}              


\definecolor{iccvblue}{rgb}{0.21,0.49,0.74}
\usepackage[pagebackref,breaklinks,colorlinks,allcolors=iccvblue]{hyperref}

\usepackage{url}
\usepackage{algorithm}
\usepackage{algorithmic}
\usepackage{multirow}%
\usepackage{amsfonts}%
\usepackage{amsthm}%
\usepackage{mathrsfs}%
\usepackage[title]{appendix}%
\usepackage{xcolor}%
\usepackage{textcomp}%
\usepackage{manyfoot}%
\usepackage{booktabs}%
\usepackage{listings}%
\usepackage{anyfontsize}
\usepackage{color}
\usepackage{bbding}
\usepackage{longtable}
\usepackage{supertabular}
\usepackage{colortbl}
\usepackage{tocloft}
\usepackage{wrapfig} 
\usepackage{mathrsfs}
\usepackage{pifont} 
\usepackage{graphicx}

\def\paperID{1161} 
\def\confName{ICCV}
\def\confYear{2025}

\title{Generalized Tensor-based Parameter-Efficient Fine-Tuning via Lie Group Transformations}

\author{Chongjie Si$^1$, Zhiyi Shi$^2$, Xuehui Wang$^1$, Yichen Xiao$^3$, Xiaokang Yang$^1$, Wei Shen$^1$\textsuperscript{\ding{41}}\\
$^1$MoE Key Lab of Artificial Intelligence, AI Institute, Shanghai Jiao Tong University.\\ $^2$Harvard University. $^3$Southeast University.\\
{\tt\small chongjiesi@sjtu.edu.cn, wei.shen@sjtu.edu.cn}
}

\begin{document}
\maketitle
\input{sec/0.abstract}    
\input{sec/1.intro}
\input{sec/2.related}
\input{sec/3.preliminary}
\input{sec/4.method}

\input{sec/5.experiment}

\input{sec/6.conclusion}

{
    \small
    \bibliographystyle{ieeenat_fullname}
    \bibliography{main}
}

\newpage

\input{sec/supp}

\end{document}

%% file: sec/0.abstract.tex
\begin{abstract}

Adapting pre-trained foundation models for diverse downstream tasks is a core practice in artificial intelligence.
However, the wide range of tasks and high computational costs make full fine-tuning impractical. 
To overcome this, parameter-efficient fine-tuning (PEFT) methods like LoRA have emerged and are becoming a growing research focus.
Despite the success of these methods, they are primarily designed for linear layers, focusing on two-dimensional matrices while largely ignoring higher-dimensional parameter spaces like convolutional kernels.
Moreover, directly applying these methods to higher-dimensional parameter spaces often disrupts their structural relationships. 
Given the rapid advancements in matrix-based PEFT methods, rather than designing a specialized strategy, we propose a generalization that extends matrix-based PEFT methods to higher-dimensional parameter spaces without compromising their structural properties.
Specifically, we treat parameters as elements of a Lie group, with updates modeled as perturbations in the corresponding Lie algebra. 
These perturbations are mapped back to the Lie group through the exponential map, ensuring smooth, consistent updates that preserve the inherent structure of the parameter space.
Extensive experiments on computer vision and natural language processing validate the effectiveness and versatility of our approach, demonstrating clear improvements over existing methods.
Codes are available at \url{https://github.com/Chongjie-Si/Subspace-Tuning}.

\end{abstract}

%% file: sec/1.intro.tex
\section{Introduction}

Recent progress in large foundation models (LFMs) \cite{qin2023chatgpt, touvron2023llama, touvron2023llama2, llama3modelcard} has demonstrated their remarkable efficacy across a broad spectrum of domains, including computer vision \cite{si2025maintaining} and natural language processing \cite{wang2018glue}. 
Nonetheless, fully fine-tuning these models—which often comprise hundreds of millions to hundreds of billions of parameters—exacts a significant computational and memory cost \cite{ma2024segment, raffel2020exploring}. 
Such extensive resource demands substantially hinder the practical deployment of LFMs in diverse applications.

To alleviate these challenges, recent research has increasingly focused on parameter-efficient fine-tuning (PEFT) approaches \cite{ hu2021lora, liu2024dora, zaken2021bitfit, si2024see, si2025unleashing}, which facilitate the adaptation of LFMs with minimal computational overhead and reduced GPU memory consumption. 
These techniques selectively optimize only a limited subset of the model’s parameters, thereby preserving the original architecture while still achieving commendable task-specific performance. 
Among these methods, low-rank adaptation (LoRA) \cite{hu2021lora} is particularly notable. 
Building on the notion that fine-tuning can be interpreted as navigating a ``low-dimensional manifold'' \cite{aghajanyan2020intrinsic, li2018measuring}, LoRA posits that modifications to a weight matrix $\mathbf{W}\in\mathbb{R}^{n\times m}$ can be effectively represented by a low-rank update. 
Specifically, it decomposes the update $\Delta\mathbf{W}$ into the product of two smaller matrices, $\mathbf{A}\in\mathbb{R}^{n\times r}$ and $\mathbf{B}\in\mathbb{R}^{r\times m}$, with $r\ll \min\{n, m\}$.
The update is incorporated by adding $\Delta\mathbf{W}$ to the pre-trained weights, while only the low-rank factors $\mathbf{A}$ and $\mathbf{B}$ are optimized during training. 
Following the advent of LoRA, LoRA has inspired a host of weight update strategies, leading to the development of numerous LoRA variants that extend and refine its core principles \cite{liu2024dora,si2025maintaining, meng2024pissa, wang2024milora, wang2024loraga, bershatsky2024lotr}.

Notably, the majority of existing PEFT approaches have been tailored for two-dimensional matrices (i.e., linear layers), largely overlooking other high-dimensional parameters like convolutional layers, which are inherently four-dimensional tensors.
However, many modern foundation models—particularly in the visual domain—rely heavily on high-dimensional operations, such as convolutional layers in ConvNeXt \cite{woo2023convnext} and Stable Diffusion \cite{rombach2022highstablediffusion}.
Moreover, adapting current PEFT methods to these higher-dimensional parameter spaces is non-trivial, as it may disrupt the inherent structural relationships within these tensors, such as the spatial locality of convolutional kernels.

While a bespoke PEFT strategy for each high-dimensional parameter space could address these issues, such an approach would quickly become impractical due to the predominance of linear weights in most foundation models and the probable diversity of parameter structures across different architectures.
Instead, it is more desirable to explore whether matrix-based PEFT methods, originally formulated for two-dimensional linear layers, can be generalized to higher-dimensional parameter spaces without compromising their unique structural properties. Such a generalization would allow the community to build on the rapid advancements in existing PEFT techniques while extending their applicability to a wilder range of model architectures, ultimately streamlining the adaptation process and enhancing flexibility across different paradigms.

In this work, we aim to extend existing matrix-based PEFT methods to higher-dimensional parameter spaces while preserving their inherent structural properties.
Our approach is founded on the observation that the changes in weights after fine-tuning are typically very small \cite{si2025unveiling} and nonzero\footnote{This also aligns with the practical consideration that, parameters of neural networks are rarely exactly zero due to the finite precision of floating-point representations in digital computers.}, thereby constituting small perturbations to the pre-trained weights \cite{du2025loca}.
Under these conditions, such parameters form a smooth manifold. 
When endowed with a suitably defined multiplication, these parameters can be interpreted as elements of a Lie group \cite{chevalley2018theory}.
This framework enables us to represent small perturbations $\Delta\mathbf{W}$ within the corresponding Lie algebra, a linear vector space where classical operations hold. 
By leveraging the local diffeomorphism property of the exponential map from the Lie algebra to the Lie group, we execute parameter updates that effectuate smooth transitions while preserving their local structural correlations. 
Consequently, our method ensures that the updates remain faithful to the underlying manifold structure, thereby amalgamating the advantages of matrix-based PEFT techniques and preservation of structure property of parameter space within the framework of Lie groups.
Extensive experimental evidence substantiates the superiority of our method. 
We hope that this approach will further catalyze innovative developments in the field of PEFT.

%% file: sec/2.related.tex
\section{Background}

\subsection{Parameter Efficient Fine-tuning}
Parameter-Efficient Fine-tuning (PEFT) addresses the computational and memory challenges of adapting large models by tuning a small subset of parameters, achieving competitive performance with reduced resource demands \cite{ding2023parameter, si2024see}. 
PEFT methods mainly fall into three categories: adapter-based, prompt-based, and low-rank adaptation \cite{liu2024dora, si2024see, si2025weight}.

Adapter-based methods \cite{houlsby2019parameter, chen2022adaptformer} insert lightweight modules into existing layers for task-specific adaptation, while prompt-based techniques \cite{lester2021power, razdaibiedina2023residual} add learnable tokens to the input to guide the model. 
Low-rank adaptation, introduced by LoRA \cite{hu2021lora}, models weight updates through low-rank adaptation, enabling efficient integration with pre-trained parameters. 
Recent advances \cite{liu2024dora, si2025unleashing, si2025map} further enhance its scalability and efficiency, establishing low-rank adaptation as a key approach for future PEFT developments.

\subsection{PEFT for High-dimensional Layers}

Despite the rapid development of PEFT, most existing methods are designed for linear layers, focusing primarily on matrices, with limited efforts targeting higher-dimensional parameters.
Considering that the highest parameter dimension in current foundation models is typically a four-dimensional tensor, represented by convolutional kernels, it serves as an ideal example for our study. While we focus on convolutional layers in this work, the proposed method is designed with scalability in mind, leaving open the possibility of applying it to even higher-dimensional parameter spaces in future models.
Taking convolutional layers as examples, to the best of our knowledge, only two methods—FLoRA \cite{si2025maintaining} and Atom-filter \cite{chen2024large}—specifically address convolutional fine-tuning. 
Specifically, FLoRA leverages Tucker decomposition to learn a low-rank 4D core tensor along with projection matrices for each dimension to model convolutional updates.
In contrast, Atom-filter decomposes the convolution operation into filter atoms and corresponding coefficients, updating the convolution through these components.


\subsection{Lie Group Theory}
Lie groups are continuous groups that combine algebraic structures with smooth manifold properties, allowing for group operations (multiplication and inversion) to be differentiable. 
They are widely used in mathematics and physics to model continuous transformations, such as rotations and scaling. 
Associated with every Lie group is a corresponding Lie algebra, which serves as its linearized tangent space at the identity element. 
The Lie algebra provides a simpler, linear structure where operations are easier to perform, and through the exponential map, elements in the Lie algebra can be mapped back to the Lie group. 
This relationship ensures that small perturbations in the Lie algebra result in smooth transformations in the Lie group.

In machine learning, Lie groups have been used in optimization on manifolds \cite{leake2021optimization}, ensuring parameter updates respect intrinsic geometric structures.
Applications include orthogonal parameter updates \cite{plumbley2004lie} and modeling rotations or transformations in computer vision \cite{liu2023lie,shutty2023computing}.
However, most prior work focuses on linear transformations or specific geometric tasks, without addressing high-dimensional parameters, which also form a smooth parameter space.

%% file: sec/3.preliminary.tex
\section{Preliminary}

\subsection{Low-rank Adaptation}
LoRA \cite{hu2021lora} models the weight updates of a pre-trained weight matrix $\mathbf{W} \in \mathbb{R}^{n \times m}$ using a low-rank decomposition form of $\Delta\mathbf{W} = \mathbf{A}\mathbf{B}$, where $\mathbf{A} \in \mathbb{R}^{n \times r}$ and $\mathbf{B} \in \mathbb{R}^{r \times m}$, with $r \ll \min(n,m)$. 
During the forward pass, the original computation $\mathbf{h} = \mathbf{W} \mathbf{x}$ is modified as
\begin{equation}
\mathbf{h} = \mathbf{W} \mathbf{x} + \Delta\mathbf{W}\mathbf{x} = \mathbf{W} \mathbf{x} + \mathbf{A}\mathbf{B} \mathbf{x}.
\end{equation}
In practice, $\mathbf{A}$ is initialized with a Gaussian distribution, while $\mathbf{B}$ is set to zeros, ensuring that $\Delta\mathbf{W}$ is initially zero.

\subsection{Matrix-based PEFT Methods for High-dimensional Layers}\label{sec pre convolution}

Although existing matrix-based PEFT methods can be adapted to convolutional layers, they often disrupt the spatial locality of convolutional kernels.
Taking LoRA as an example, when adapting LoRA for convolutional layers with the weights $\mathcal{W} \in \mathbb{R}^{d_{\mathrm{in}} \times d_{\mathrm{out}} \times k \times k}$. 
Here, $d_{\mathrm{in}}$ and $d_{\mathrm{out}}$ represent the input and output channel dimensions, and $k$ denotes the kernel size. 
LoRA reshapes the matrix updates into a four-dimensional kernel tensor:
\begin{equation}
\mathcal{W} \rightarrow \mathcal{W} + \Delta\mathcal{W} = \mathcal{W} + \mathrm{Reshape}(\Delta\mathbf{W}, \mathcal{W}).
\end{equation}
The function $\mathrm{Reshape}(\Delta\mathbf{W}, \mathcal{W})$ transforms the low-rank matrix $\Delta\mathbf{W}$ into the original four-dimensional shape of $\mathcal{W}$.
However, this reshaping process disrupts the inherent local structure in convolution. 
Adjacent elements within a convolutional kernel often originate from distant rows and columns in the reshaped matrix, breaking the spatial locality. 
This loss of locality compromises the inherent spatial correlations within the kernel, which are crucial for capturing local patterns in visual tasks, as shown in \cite{si2025maintaining}.

%% file: sec/4.method.tex
\section{Method}

\begin{figure*}
    \centering
    \includegraphics[width=\linewidth]{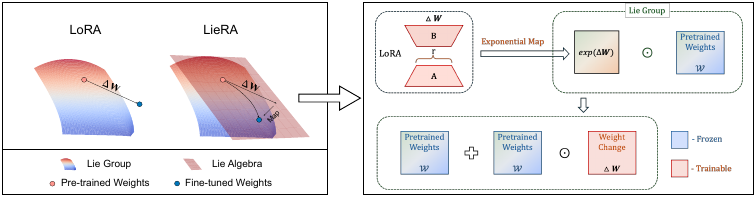}
    \caption{Illustration comparing LoRA and LoRA in our framework, LieRA. Parameters with the same structure are considered elements of a Lie group. LoRA does not account for the inherent structural properties of higher-dimensional parameters, leading to a loss of their structure (i.e., the updates are not within the Lie group). In contrast, LieRA simulates updates at the Lie algebra level and then uses the exponential map to map these updates back into the Lie group, preserving the inherent structure of the parameters during the weight update process. Specifically, we represent small perturbations $\Delta\mathbf{W}$ within the Lie algebra and perform updates using the exponential map to transition smoothly into the Lie group. The first-order Taylor approximation is then applied to simplify the exponential map for efficient computation, enabling matrix-based PEFT updates while maintaining the structural properties of the parameters.}
    \label{fig:framework}
\end{figure*}

In this section, we present our method (Fig. \ref{fig:framework}) to adapt matrix-based approaches to high-dimensional parameters while preserving their structure property. 
To achieve this, we leverage the smooth manifold structure and continuous transformation properties of Lie groups, which ensure that updates remain consistent with the intrinsic spatial locality of the kernel. 
We begin by constructing a Lie group for high-dimensional parameters and demonstrating that they can be treated as elements of this group.
Considering the practical scenarios of existing LFMs, we focus solely on convolution as a representative case for high-dimensional parameters.

\subsection{Lie Groups and Lie Algebras for Convolution}

Considering the properties of convolutional kernel parameters in existing neural networks, we construct the set $G = \Big\{\mathcal{W} \in \mathbb{R}^{C_{\mathrm{in}} \times C_{\mathrm{out}} \times k \times k} \,\Big|\, W_{c,i,j,l} \neq 0,\; \forall\, c,i,j,l \Big\}$, and treat convolutional kernel parameters as elements of $G$.
With an appropriately defined group operation $\circ$, $(G, \circ)$ can form a valid group.
It is evident that common binary operations for updating convolutional kernels, such as addition (used in most matrix-based PEFT methods), tensor multiplication, or convolution itself, fail to satisfy the group axioms\footnote{Addition lacks an identity element, and tensor multiplication or convolution does not guarantee an inverse for every element.}.
After careful consideration, we define a binary operation $\odot$ on $G$ via element-wise multiplication (i.e., the Hadamard product) as the group operation. 
This choice ensures that $\odot$ is closed in $G$, associative, and admits the identity element $\mathcal{I}$ ($\mathcal{I}$ denotes a tensor with all entries equal to one). 
Furthermore, every element $\mathcal{W} \in G$ has an inverse $\mathcal{W}^{-1}$, defined element-wise as $(\mathcal{W}^{-1}){c,i,j,l} = 1/W{c,i,j,l}$. 
This construction guarantees that $(G, \odot)$ forms a valid group.

Next, we demonstrate that $G$ is endowed with the structure of a smooth manifold.
Observing that
\begin{equation}
    G \cong \prod_{c,i,j,l} (\mathbb{R} \setminus \{0\}),
\end{equation}
and recalling that $\mathbb{R} \setminus \{0\}$ constitutes a one-dimensional Lie group under multiplication, it follows that $G$ is a finite Cartesian product of these one-dimensional Lie groups.
Thus, $G$ inherits a smooth manifold structure and forms an Abelian Lie group of dimension
$N = C_{\mathrm{out}} \cdot C_{\mathrm{in}} \cdot k^2$.

The corresponding Lie algebra, denoted by $\mathfrak{g}$, is naturally isomorphic to $\mathbb{R}^{C_{\mathrm{out}} \times C_{\mathrm{in}} \times k \times k}$, which provides a direct and convenient representation of convolutional kernel parameters in a linear space.
Operations in $\mathfrak{g}$ are defined by element-wise addition and scalar multiplication, making it a linear vector space where updates can be performed easily and efficiently. 
This simplifies optimization compared to working directly in the nonlinear Lie group $G$.
Since $G$ is an Abelian group, the Lie bracket on $\mathfrak{g}$ is trivial, meaning that all elements commute.

A crucial component in this framework is the exponential map, which provides a bridge between the linear Lie algebra $\mathfrak{g}$ and the nonlinear Lie group $G$.
For any $\Delta \mathbf{A} \in \mathfrak{g}$, the exponential map is computed element-wise as:
\begin{equation}
    (\exp(\Delta A))_{c,i,j,l} = \exp\big((\Delta A)_{c,i,j,l}\big).
\end{equation}
The exponential map serves as a local diffeomorphism, ensuring that small perturbations in $\mathfrak{g}$ correspond to smooth, structure-preserving transformations in $G$, thereby ensuring that  parameter updates remain continuous and preserve the spatial locality intrinsic to the convolutional kernels.

\subsection{Lie Group-Based Parameter Update}

Let $\mathcal{W} \in G$ denote the pre-trained convolution kernel, and let $\Delta \mathcal{W} \in \mathfrak{g}$ represent a perturbation residing in the Lie algebra. 
Rather than performing a conventional additive update $\mathcal{W} \rightarrow \mathcal{W} + \Delta \mathcal{W}$ as in most existing methods, we propose a multiplicative update using the group operation:
\begin{equation}
   \mathcal{W} \rightarrow \mathcal{W} \odot \exp(\Delta \mathcal{W}),
\end{equation}
where $\exp(\Delta \mathcal{W})$ is computed element-wise through the exponential map.

The choice of this multiplicative update is crucial for preserving the geometric structure of the parameter space.
Intuitively, the multiplicative update scales each element proportionally to its current value, preserving the relative structure of the kernel and maintaining spatial locality.
Additionally, since $G$ is closed under the group operation $\odot$, if both $\mathcal{W} \in G$ and $\exp(\Delta \mathcal{W}) \in G$, then the updated kernel $\mathcal{W} \odot \exp(\Delta \mathcal{W})$ remains within $G$.
This guarantees that the updated parameters respect the group’s smooth manifold structure and ensures that they stay within a consistent and valid space throughout the fine-tuning process.

Moreover, since $\mathfrak{g}$ is a linear vector space, it allows for efficient parameterization and optimization. 
Perturbations in $\mathfrak{g}$ can be reparameterized using low-rank adaptations or other methods, significantly reducing the number of trainable parameters while remaining consistent with the Lie algebra structure.
The exponential map then ``lifts'' these compact perturbations back to the nonlinear manifold $G$, ensuring that the update remains geometrically consistent within the Lie group.

For linear layers, the Lie group-based approach can still be applied to ensure smooth and consistent updates of the weight matrix.
Specifically, the update is performed as:
\begin{equation}
    \mathbf{W}\rightarrow \mathbf{W} \odot \exp(\mathbf{\Delta\mathbf{W}}),
    \label{eq W exp deltaW}
\end{equation}
where $\exp(\Delta \mathbf{W})$ is computed element-wise. 
This formulation extends the same principle used for convolutional kernels to two-dimensional weight matrices, preserving the structure of the linear layer while ensuring stable and continuous updates.

\subsection{First-Order Taylor Approximation for the Exponential Map}

Since $\Delta \mathcal{W}$ is small, the exponential map admits a first-order Taylor expansion.
Specifically, we have
\begin{equation}
    \exp(\Delta \mathcal{W}) = \mathcal{I} + \Delta \mathcal{W} + o(\|\Delta \mathcal{W}\|),
\end{equation}
where $\mathcal{I}$ denotes the tensor with all entries equal to one, and the term $o(\|\Delta \mathcal{W}\|)$ represents a remainder that is asymptotically negligible compared to $\|\Delta \mathcal{W}\|$ as $\|\Delta \mathcal{W}\|$ to 0.
Therefore, we can approximate the exponential map as $\exp(\Delta \mathcal{W}) \approx \mathcal{I} + \Delta \mathcal{W}$.
Substituting this approximation into our multiplicative update, the update rule becomes
\begin{equation}
    \mathcal{W} \odot \exp(\Delta \mathcal{W}) \approx \mathcal{W} \odot (\mathcal{I} + \Delta \mathcal{W}) = \mathcal{W} + \mathcal{W} \odot \Delta \mathcal{W},
    \label{eq taylor}
\end{equation}
which succinctly captures the first-order effect of the $\Delta \mathcal{W}$. 
This is more computationally simplified, enabling efficient implementation while preserving the properties of the Lie group-based update.
In the subsequent experiments, we integrate our framework with LoRA based on Eq. (\ref{eq taylor}) for evaluation, and refer to the combined approach as \textit{LieRA}.

\subsection{Theoretical Insights}

Building on the previous analysis, we know that LieRA has a clear advantage when adapting to high-dimensional parameters. 
We here further show advantage of LieRA compared to LoRA in linear layers based on rank capacity \cite{zhang2024spectral}.

\input{tables/VTAB_conv}

\input{tables/Detection}

Rank capacity $\mathcal{R}$ refers to the range of possible ranks that the updated weight matrix can achieve during fine-tuning, reflecting the flexibility to capture diverse updates.
A higher rank capacity allows the model to better adapt to various tasks by leveraging more degrees of freedom. 
Specifically, for LoRA, the rank capacity is determined by the rank $r$ of the low-rank matrices, with $\mathcal{R}(\mathbf{AB}) = r$.
In contrast, due to the properties of the Hadamard product and the considering that pre-trained weights are typically nearly full-rank \cite{hu2021lora}, the rank capacity of LieRA is $R(\mathbf{W} \odot AB) = \min(n, m)$, where $n$ and $m$ are the dimensions of the original weight matrix $\mathbf{W}$. 
This results in a full-rank capacity, making LieRA far more flexible than LoRA and offering enhanced performance for downstream tasks.

%% file: tables/VTAB_conv.tex
\begin{table*}[ht]

\caption{Results of ConvNeXt-V2-B fine-tuned on the VTAB-1k benchmark for image classification.}
\label{tab:main_results}
\resizebox{\textwidth}{!}{
\begin{tabular}{lc | ccccccc | cccc | cccccccc | c}
    \toprule
    & & \multicolumn{7}{c|}{\textbf{Natural}} & \multicolumn{4}{c|}{\textbf{Specialized}} & \multicolumn{8}{c|}{\textbf{Structured}} \\
    & \rotatebox{90}{\# param (M)} & \rotatebox{90}{Cifar100} & \rotatebox{90}{Caltech101} & \rotatebox{90}{DTD} & \rotatebox{90}{Flower102} & \rotatebox{90}{Pets} & \rotatebox{90}{SVHN}  & \rotatebox{90}{Sun397} & \rotatebox{90}{Camelyon} & \rotatebox{90}{EuroSAT}   & \rotatebox{90}{Resisc45}  & \rotatebox{90}{Retinopathy} & \rotatebox{90}{Clevr-Count} & \rotatebox{90}{Clevr-Dist}  & \rotatebox{90}{DMLab} & \rotatebox{90}{KITTI-Dist}  & \rotatebox{90}{dSpr-Loc} & \rotatebox{90}{dSpr-Ori}   & \rotatebox{90}{sNORB-Azim}  & \rotatebox{90}{sNORB-Ele} & \rotatebox{90}{Average}   \\
    \midrule
    Fully FT \cite{woo2023convnext} & 102.05 & 69.0 & 91.9 & 76.1 & 99.5 & 92.1 & 89.7 & 52.5 & 86.4 & 96.0 & 88.3 & 78.4 & 93.7 & 55.9 & 56.1 & 78.4 & 96.3 & 70.2 & 39.1 & 36.3 & 78.2 \\ 

    \midrule
    LoRA$_{r=2}$ & 1.90 & 58.6 & 89.9 & 71.2 & 91.6 & 92.3 & 89.4 & 45.8 & 85.0 & 94.1 & 81.6 & 74.1 & 84.4 & 61.1 & 49.8 & 80.7 & 95.3 & 60.2 & 35.1 & 34.5 & 74.4  \\

    \rowcolor{gray!20}
    
    LieRA & 1.90 & 61.6 & 89.8 & 72.9 & 92.9 & 93.2 & 87.3 & 47.0 & 84.4 & 94.5 & 84.0 & 74.2 & 83.1 & 65.0 & 48.8 & 78.9 & 92.7 & 60.0 & 33.8 & 34.3 & 74.7  \\ \midrule

    LoRA$_{r=4}$ & 3.70 & 57.6 & 90.5 & 70.7 & 90.4 & 92.3 & 89.4 & 45.0 & 84.4 & 94.0 & 80.4 & 74.5 & 85.4 & 64.2 & 51.2 & 80.3 & 95.9 & 60.3 & 36.2 & 32.8 & 74.4  \\

    \rowcolor{gray!20}
    
    LieRA & 3.70 & 60.7 & 90.0 & 72.3 & 92.8 & 93.0 & 88.9 & 46.9 &  84.2 & 94.6 & 83.6 & 74.9 & 85.9 & 65.8 &  50.2 & 79.7 & 94.2 & 60.9 & 34.7 & 34.2 & 75.1  \\ \midrule

    LoRA$_{r=8}$ & 7.30 & 57.1 & 90.6 & 71.1 & 90.0 & 92.0 & 90.2 & 43.4 & 85.4 & 94.2 & 80.5 & 74.6 & 81.3 & 63.3 & 52.1 & 80.8 & 94.6 & 60.9 & 36.1 & 33.1 & 74.2 \\ \rowcolor{gray!20}
    
    LieRA & 7.30 & 60.8 & 90.6 & 71.3 & 92.9 & 93.1 & 89.7 & 46.9 & 85.1 & 94.4 & 83.4 & 74.8 & 86.1 & 68.9 & 50.8 & 79.9 & 94.4 & 61.3 & 35.3 & 35.7 & 75.5 \\

    \midrule

     LoRA$_{r=16}$ & 14.48 & 56.5 & 90.8 & 68.5 & 89.3 & 91.7 & 90.7 & 42.2 & 84.8 & 94.0 & 80.8 & 75.0 & 80.0 & 64.1 & 52.7 & 79.3 & 97.9 & 59.9 & 37.0 & 33.0 & 74.1  \\

     \rowcolor{gray!20}

    LieRA & 14.48 & 60.2 & 90.4 & 71.8 & 92.9 & 92.9 & 89.5 & 46.2 & 84.2 & 94.5 & 83.5 & 74.7 & 85.6 & 69.0 & 50.8 & 79.9 & 97.1 & 61.5 & 36.2 & 36.4 & 75.5 \\

    \bottomrule
    \end{tabular}}
\end{table*}

%% file: tables/Detection.tex
\begin{table*}[ht]
    \centering
\renewcommand\arraystretch{1} 
\setlength{\tabcolsep}{2.4mm}
    \caption{Results for ConvNeXt-V2-B fine-tuned on MS COCO for object detection and instance segmentation.``Base'' denotes the pre-trained backbone with its weights kept frozen.}
   \resizebox{\textwidth}{!}{
    \begin{tabular}{l | c| c c c c c c |c c c c c c | >{\columncolor{gray!10}} c}
    \toprule
         \multirow{2}{*}{\textbf{Method}} &  \multirow{2}{*}{\# \textbf{Params (M)}} & \multicolumn{6}{c|}{\textbf{Detection}}  & \multicolumn{6}{c|}{\textbf{Instance Segmentation}} & {\textbf{All}} \\
         & & mAP & AP$_{50}$ & AP$_{75}$ & AP$_{S}$ & AP$_{M}$ & AP$_{L}$ & mAP & AP$_{50}$ & AP$_{75}$ & AP$_{S}$ & AP$_{M}$ & AP$_{L}$ & Avg.\\ 
         
         \midrule

         Fully FT & 104.97 & 49.0 & 69.1 & 54.6 & 31.7 & 53.5 & 61.7 & 43.4 & 66.2 & 47.6 & 23.4 & 47.1 & 60.9 & 50.7 \\

        \midrule

        LoRA$_{r=16}$ & 17.27 & 35.5 & 55.9 & 38.9 & 22.7 & 38.4 & 45.1 & 33.6 & 53.1 & 36.3 & 17.4 & 35.9 & 48.2 & 38.4 \\ 
        \rowcolor{gray!20}

        LieRA & 17.27 & 39.1 & 59.9 & 43.1 & 25.9 & 42.6 & 49.2 & 37.0 & 57.6 & 40.2 & 20.1 & 40.0 & 53.2 & 42.3 \\

        \midrule
        
        LoRA$_{r=32}$ & 34.54 & 35.9 & 56.4 & 39.7 & 22.9 & 39.6 & 45.7 & 34.4 & 54.2 & 37.1 & 18.3 & 36.7 & 48.9 & 39.2 \\ \rowcolor{gray!20}

        LieRA & 34.54 & 40.5 & 61.2 & 45.1 & 25.3 & 44.1 & 50.8 & 38.2 & 58.9 & 41.8 & 19.8 & 41.2 & 53.9 & 43.4 \\

         \bottomrule
    \end{tabular}
    }
    \label{tab: convolution results}
\end{table*}

%% file: sec/5.experiment.tex
\section{Experiment}

\begin{figure*}[!ht]
    \centering
    \includegraphics[width=\linewidth]{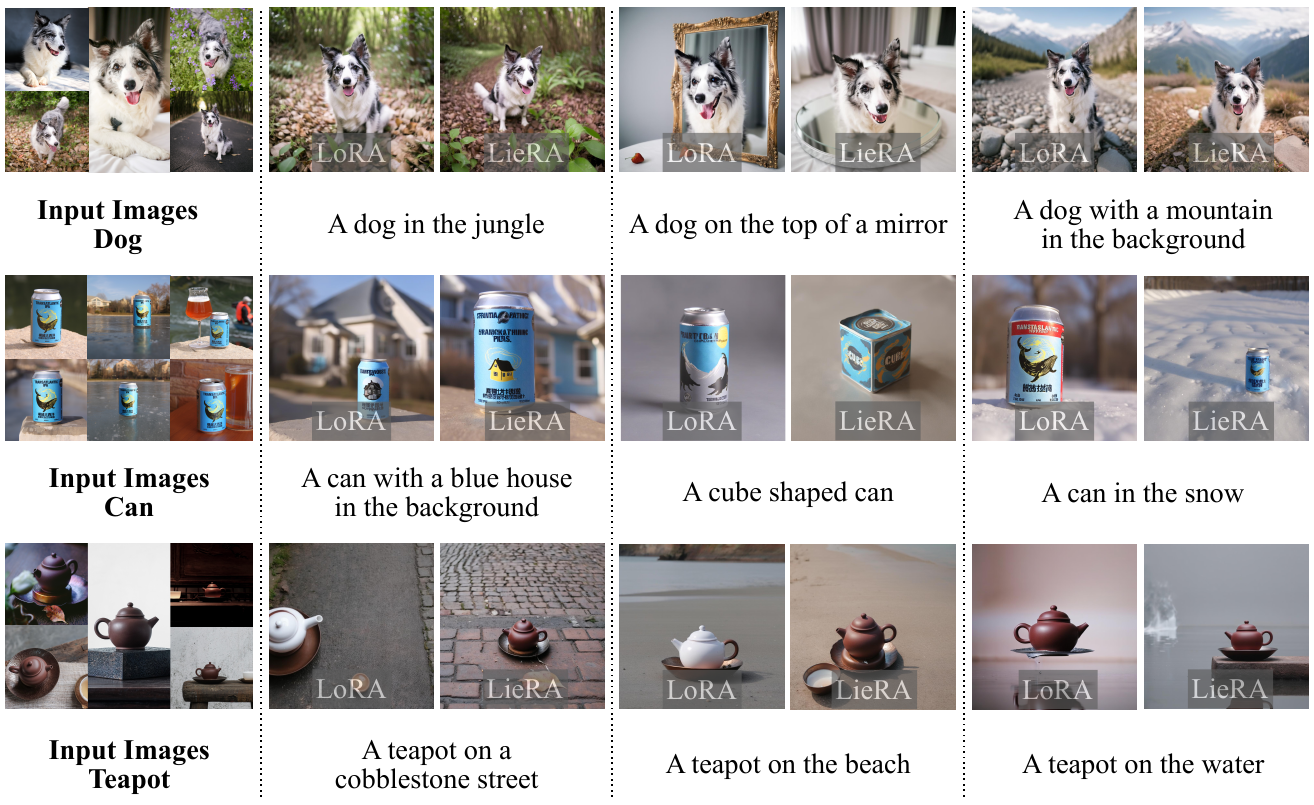}
    \caption{Comparison of images generated by stable diffusion fune-tuned by LoRA (left) and LieRA (right). Obviously, LieRA consistently produces outputs that are more accurately aligned with the subjects in the input images. Additionally, it adheres more closely to the provided prompts compared to LoRA.}
    \label{fig:stable diffusion}
\end{figure*}

\subsection{Tasks, Datasets and Models}
We perform extensive experiments spanning both computer vision (CV) and natural language processing (NLP) tasks. 
\subsubsection{Computer Vision Tasks}
Specifically, the CV tasks cover image classification, object detection and segmentation, and generative tasks.
For image classification, we fine-tune the pre-trained ConvNeXt-V2-B \cite{woo2023convnext} on the VTAB-1K benchmark to evaluate its performance across diverse datasets.
For object detection and segmentation, we fine-tune ConvNeXt-V2-B on MS COCO \cite{lin2014microsoft} using Mask R-CNN \cite{mask_rcnn} as implemented in MMDetection \cite{chen2019mmdetection}.
For the generative task, we fine-tune the SDXL5 model \cite{podell2023sdxl} to generate images that match the given prompts, aiming to produce outputs aligned with the learned subject matter.

\subsubsection{Natural Language Processing Tasks}
For NLP tasks, we evaluate our method on commonsense reasoning and natural language understanding (NLU) tasks.
Specifically, following \cite{liu2024dora,si2025unleashing}, we fine-tune LLaMA-7B \cite{touvron2023llama}, and LLaMA3-8B \cite{llama3modelcard} on commonsense reasoning datasets.
For NLU task, we fine-tune DeBERTaV3-base \cite{he2021debertav3} on the General Language Understanding Evaluation (GLUE) benchmark \cite{wang2018glue}.

\subsection{Implementation Details}

We primarily integrate LoRA for evaluation with varying parameter budgets.
The rank $r$ for both LoRA and LieRA is chosen from \{2, 4, 8, 16, 32\}.
Other hyper-parameters are initialized according to the original papers. 
For computer vision tasks, we perform fine-tuning on all convolutional layers of ConvNeXt-V2-B, and all layers of SDXL. 
In the case of natural language processing tasks, we fine-tune all linear layers across all models, as outlined in \cite{si2025unleashing,liu2024dora}. 
We use the publicly available PyTorch implementation \cite{paszke2019pytorch} to carry out all baseline comparisons, with all experiments conducted on NVIDIA A100 GPUs. More details on training are shown in the Appendix.

\subsection{Results on Computer Vision Tasks}

Table \ref{tab:main_results} shows the results of fine-tuning ConvNeXt-V2-B on the VTAB-1K benchmark, comparing the performance of different LoRA configurations and the LieRA method across multiple natural, specialized, and structured tasks. 
For natural category tasks, the performance differences between LoRA and LieRA are quite significant across different rank settings.
Under identical parameter configurations, LieRA consistently outperforms LoRA, particularly in tasks like SVHN and Sun397. 
Since LieRA effectively preserves the inherent structural properties of convolutional kernels, maintaining spatial locality and multi-scale correlations during the learning process, it demonstrates an enhanced ability to capture and leverage task-specific features, resulting in superior performance across most tasks.

Additionally, table \ref{tab: convolution results} presents the results of object detection and instance segmentation tasks on the MS COCO dataset.
The experimental results clearly validate the superiority of LieRA, demonstrating its effective preservation of high-dimensional parameter space structures.

Additionally, Fig. \ref{fig:stable diffusion} illustrates the results of the generation experiment. 
The goal of this experiment was to ensure that the generated images are consistent with both the input images and the text prompt. 
Clearly, the images generated by LieRA exhibit superior fidelity, aligning better with the subjects of the input images compared to those generated by standard LoRA.
For instance, the images generated by LoRA of a can show significant deviations from the input, whereas the outputs of LieRA maintained high consistency with the original images.

\input{tables/CR}
\input{tables/NLU}

Overall, the results from the CV experiments validate the outstanding performance of LieRA in adapting to convolutional layers. 
By integrating LoRA’s updates within the Lie group structure, LieRA effectively preserves the structural properties of high-dimensional parameter spaces, leading to significant improvements when adapting to such spaces.

\subsection{Results on Natural Language Processing Tasks}

Table \ref{tab:results of commonsense} presents the results of fine-tuning LLaMA-7B and LLaMA3-8B on commonsense reasoning tasks. The results clearly demonstrate that LieRA consistently outperforms LoRA, indicating its superior capability in adapting to task-specific representations. Specifically, when fine-tuning LLaMA-7B, LieRA achieves an average performance improvement of 4.3\% over LoRA across all tasks, highlighting its enhanced expressiveness in modeling complex relationships. Notably, LieRA provides substantial gains on tasks such as WinoGrande (+25.7\%) and ARC-e (+4.3\%), where capturing intricate reasoning patterns is crucial. This suggests that LieRA is more effective in capturing fine-grained dependencies, particularly in tasks requiring nuanced contextual understanding.
A similar trend is also observed for LLaMA3-8B, where LieRA surpasses LoRA across all tested benchmarks. The improvement is particularly evident in high-complexity tasks such as HellaSwag and ARC-c, where LieRA exhibits strong generalization capabilities. These results confirm that LieRA provides a more expressive adaptation mechanism for large-scale language models, effectively capturing key task-specific patterns while maintaining parameter efficiency.

\begin{table*}[!ht]
     \centering
    \renewcommand\arraystretch{1} 
    \setlength{\tabcolsep}{3mm}
    \caption {Training time (minutes/epoch) and GPU usage (GB) of LieRA compared to LoRA.}
    \resizebox{\textwidth}{!}{
    \begin{tabular}{c |c | c c c c | c |c c c c c c }
    \toprule
         \multirow{3}{*}{\textbf{Method}} & \multicolumn{5}{c|}{\textbf{ConvNeXt-V2-B}} & \multicolumn{7}{c}{\textbf{DeBERTaV3-base}}\\ \cline{2-13}
         & \multirow{2}{*}{ \textbf{Rank}} & \multicolumn{2}{c}{\textbf{COCO}} & \multicolumn{2}{c|}{\textbf{Cifar100}} & \multirow{2}{*}{\textbf{Rank}} & \multicolumn{2}{c}{\textbf{CoLA}} & \multicolumn{2}{c}{\textbf{SST-2}} &  \multicolumn{2}{c}{\textbf{STS-B}}\\
         & & Time & GPU & Time & GPU & & Time & GPU & Time & GPU & Time & GPU \\
         
         \midrule
         
        LoRA & $r=16$ & 55.42 & 8.62 & 0.13 & 9.14 & $r=2$ & 0.52 & 5.07 & 6.45 & 6.85 & 0.56 & 6.85 \\

        LieRA & $r=16$ & 50.17 & 9.97 & 0.13 & 9.14 & $r=2$ & 0.63 & 5.41 & 8.33 & 7.19 & 0.70 & 7.19 \\
        
        \bottomrule
    \end{tabular}
    }
    \label{tab: training time}
\end{table*}

Table \ref{tab: deberta results} further validates these findings by presenting the fine-tuning results of DeBERTaV3 on the GLUE development set. Consistent with the commonsense reasoning results, LieRA achieves superior performance across almost all benchmarks, significantly outperforming LoRA at equivalent parameter budgets. Notably, on DeBERTaV3-Base, LieRA surpasses LoRA by 0.84\% on average, with notable gains in CoLA (+1.6\%), MRPC (+0.99\%), and STS-B (+1.06\%), which suggests that LieRA is particularly effective in tasks requiring nuanced linguistic understanding and semantic similarity assessments. Similarly, for DeBERTaV3-Large, LieRA continues to outperform LoRA, achieving a notable improvement on CoLA (+1.3\%), indicating enhanced robustness in capturing grammatical acceptability.
Overall, the results from the NLP experiments clearly demonstrate the superior performance of LieRA adapted the linear layers.

\subsection{Ablation Study}

\subsubsection{Approximation of First-Order Taylor Expansion}

\begin{table}[!ht]
     \centering
    \renewcommand\arraystretch{1} 
    \setlength{\tabcolsep}{3mm}
    \caption {The performance and training resource requirements of LieRA with and without the first-order Taylor approximation. The units for training time and GPU memory usage are minutes/epoch and GB, respectively. ``TA'' represents for Taylor Approximation, and ``Avg.'' denotes the average performance.}
    \resizebox{\linewidth}{!}{
    \begin{tabular}{c |c c c | c c c }
    \toprule
         \multirow{2}{*}{\textbf{Method}} & \multicolumn{3}{c|}{\textbf{VTAB-1k} ($r=16$)} & \multicolumn{3}{c}{\textbf{COCO} ($r=16$)}\\ \cline{2-7}
         & Avg. & Time & GPU & Avg. & Time & GPU \\
         
         \midrule
         
        with TA & 75.5 & 0.13 & 9.14 & 42.3 & 50.17 & 9.97 \\

        without TA & 75.7 & 0.14 & 9.18 & 42.7 & 76.28 & 14.74 \\
        
        \bottomrule
    \end{tabular}
    }
    \label{tab: first order taylor}
\end{table}

We evaluate the impact of using a first-order Taylor expansion approximation.
We compare the performance and training resource requirements of fine-tuning ConvNeXT-V2-B with and without the approximation (i.e., LieRA in Eq. (\ref{eq taylor}) vs. the non-approximate version in Eq. (\ref{eq W exp deltaW})).
The results, as shown in Table, indicate that the performance of the non-approximate LieRA is slightly better than that of LoRA with the first-order Taylor expansion. 
This subtle difference is primarily due to the small magnitude of $\Delta \mathbf{W}$, which can be considered a perturbation, leading to minimal error from the Taylor expansion.
However, the approximate LieRA requires much fewer training resources, with negligible performance loss, making it a more efficient choice.

\subsubsection{Training Costs}
We evaluate the training costs across different configurations, ensuring that all hyper-parameters, including batch size and number of epochs, remain the same. 
The results in Table \ref{tab: training time} show that LieRA and LoRA require similar GPU resources and training time, with LieRA even outperforming LoRA in terms of training time for convolutional layers. 
When adapting matrix-based methods, our approach demands comparable training resources to existing methods, while achieving significant improvements in performance, highlighting the advantages of our framework.

\subsubsection{Coupling with Other Methods}

Table \ref{tab: extending} reports the performance of PISSA \cite{meng2024pissa} and DoRA \cite{liu2024dora} on computer vision tasks (VTAB and COCO), both with and without our framework. 
Coupling our framework with either method consistently improves performance. 
For example, PISSA+Ours increases the VTAB score from 74.7 to 75.7 and the COCO score from 38.2 to 42.4, raising the average from 56.5 to 59.1. 
Similar improvements are observed for DoRA. These results affirm the efficacy of our framework in enhancing matrix-PEFT methods to higher dimensional parameter spaces.

\begin{table}[!ht]
     \centering
    \renewcommand\arraystretch{1} 
    \setlength{\tabcolsep}{6mm}
    \caption {The performance of PISSA and DoRA coupled with ours on computer vision tasks.}
    \resizebox{\linewidth}{!}{
    \begin{tabular}{c |c c c }
    \toprule
         \textbf{Benchmark (Avg.)} & VTAB & COCO & Avg. \\ \midrule {PISSA}$_{r=16}$ & 74.7 & 38.2 & 56.5 \\
         {PISSA+Ours} & 75.7 & 42.4 & 59.1 \\ \midrule
         {DoRA}$_{r=16}$ & 74.7 & 38.4 & 56.6
         \\
         {DoRA+Ours} & 75.5 & 42.5 & 59.0 \\              
        \bottomrule
    \end{tabular}
    }
    \label{tab: extending}
\end{table}

%% file: tables/CR.tex
\begin{table*}[!ht]
 \renewcommand\arraystretch{1.0}
 \setlength{\tabcolsep}{2.4mm}
    \centering
    \caption{Results for LLaMA-7B and LLaMA3-8B fine-tuned on commonsense reasoning tasks.}
    \resizebox{\textwidth}{!}{
    \begin{tabular}{ l c | c c c  c c c c c | l }
    \toprule
     \textbf{Method} & \textbf{Params(\%)} & \textbf{BoolQ} & \textbf{PIQA} & \textbf{SIQA} & \textbf{HellaS.} & \textbf{WinoG.} & \textbf{ARC-e} & \textbf{ARC-c} & \textbf{OBQA} & \multicolumn{1}{c}{\textbf{Avg.}} \\
    \toprule
    ChatGPT  & - & 73.1 & 85.4 & 68.5 & 78.5 & 66.1 & 89.8 & 79.9 & 74.8 & 77.0 \\ \midrule
    
    \multicolumn{11}{c}{\textit{Fine-tuning LLaMA-7B}} \\ \midrule
    
    LoRA$_{r=16}$ & 0.42\% & 69.9 & 77.8 & 75.1 & 72.1 & 55.8 & 77.1 & 62.2 & 78.0 & 70.9 \\ 
    
    LieRA & 0.42\% & 68.5 & 80.7 & 78.1 & 75.6 & 75.5 & 80.8 & 62.7 & 79.4 & 75.2 \\ \hline 

    Series \cite{houlsby2019parameter} & 0.99\% & 63.0 & 79.2 & 76.3 & 67.9 & 75.7 & 74.5 & 57.1 & 72.4 & 70.8 \\

    Parallel \cite{he2021towards} & 3.54\% & 67.9 & 76.4 & 78.8 & 69.8 & 78.9 & 73.7 & 57.3 & 75.2 & 72.2 \\

     LoRA$_{r=32}$ & 0.83\% & 68.9 & 80.7 & 77.4 & 78.1 & 78.8 & 77.8 & 61.3 & 74.8 & 74.7 \\ 
    
    \rowcolor{gray!20}

    LieRA & 0.83\% & 68.2 & 80.3 & 76.8 & 79.0 & 81.5 & 81.4 & 64.1 & 78.4 & 76.2\\

    \midrule
    
     \multicolumn{11}{c}{\textit{Fine-tuning LLaMA3-8B}} \\ 
     
     \midrule

    LoRA$_{r=16}$ & 0.35\% & 72.3 & 86.7 & 79.3 & 93.5 & 84.8 & 87.7 & 75.7 & 82.8 & 82.8 \\

    LieRA & 0.35\% & 74.7 & 87.9 & 79.8 & 95.6 & 84.6 & 91.4 & 79.9 & 86.8 & 85.1  \\
    
    \hline

    PISSA \cite{meng2024pissa} & 0.70\% & 67.1 & 81.1 & 77.2 & 83.6 & 78.9 & 77.7 & 63.2 & 74.6 & 75.4\\

    MiLoRA \cite{wang2024milora} & 0.70\% & 68.8 & 86.7 & 77.2 & 92.9 & 85.6 & 86.8 & 75.5 & 81.8 & 81.9 \\

    LoRA$_{r=32}$ & 0.70\% & 70.8 & 85.2 & 79.9 & 91.7 & 84.3 & 84.2 & 71.2 & 79.0 & 80.8 \\ 
    
    \rowcolor{gray!20}

    LieRA & 0.70\% & 74.3 & 88.7 & 81.3 & 95.4 & 85.5 & 90.5 & 80.3 & 86.2 & 85.3  \\
    
    \bottomrule
    \end{tabular}}
    \label{tab:results of commonsense}
\end{table*}

%% file: tables/NLU.tex
\begin{table*}[ht]
    \centering
    \renewcommand\arraystretch{1} 
    \setlength{\tabcolsep}{2.8mm}
    \caption {Results for DeBERTaV3 fine-tuned on GLUE development set. ``FT'' refers to fully fine-tuning, while ``Base' and ``Large'' denote DeBERTaV3-base and DeBERTaV3-large, respectively.}
    \resizebox{\textwidth}{!}{
    \begin{tabular}{l | c| c c c c c c c c |>{\columncolor{gray!10}}c}
    \toprule
         \multirow{2}{*}{\textbf{Method}} &  \multirow{2}{*}{\textbf{Params(\%)}} & \textbf{MNLI} & \textbf{SST-2} &\textbf{CoLA} & \textbf{QQP} & \textbf{QNLI} & \textbf{RTE} & \textbf{MRPC} & \textbf{STS-B} & \textbf{All}\\
         & & Acc & Acc & Mcc & Acc & Acc & Acc & Acc & Corr & Avg. \\ 
         \midrule
        
        Base(FT) & 100\% & 89.90 & 95.63 & 69.19 & 91.87 & 94.03 & 83.75 & 90.20 & 91.60 & 88.27 \\ \midrule
        
         LoRA$_{r=2}$ & 0.18\% & 90.03 & 93.92 & 69.15 & 90.61 & 93.37 &  87.01 & 90.19 & 90.75 & 88.13  \\  \rowcolor{gray!20}

         LieRA & 0.18\% & 90.02 & 95.41 & 70.75 & 91.27 & 93.92 & 87.36 & 91.18 & 91.81 & 88.97 \\  \midrule

        LoRA$_{r=8}$ & 0.72\% & 89.80 & 93.69 & 69.30 & 91.78 & 92.97 & 86.28 & 90.68 & 91.62 & 88.27  \\  \rowcolor{gray!20}

         LieRA & 0.72\% & 90.80 & 95.87 & 70.57 & 92.11 & 93.15 & 87.36 & 91.67 & 91.15 & 89.09\\   \bottomrule \toprule

         Large(FT) & 100\% & 91.81 & 96.93 & 75.27 & 93.01 & 96.02 & 92.68 & 92.20 & 92.98 & 91.36 \\ \midrule
        
         LoRA$_{r=2}$ & 0.20\% & 91.33 & 95.87 & 73.89 & 91.84 & 95.14 & 91.69 & 90.68 & 92.85 & 90.41 \\  \rowcolor{gray!20}

         LieRA & 0.20\% & 91.54 & 96.67 & 74.51 & 92.65 & 95.30 & 92.06 & 92.40 & 92.97 & 91.01 \\  \midrule

        LoRA$_{r=8}$ & 0.80\% & 91.38 & 96.33 & 74.48 & 92.54 & 95.48 & 92.05 & 91.17 & 92.92 & 90.79 \\  \rowcolor{gray!20}

         LieRA  & 0.80\% & 91.50 & 96.10 & 75.78 & 92.65 & 95.61 & 92.42 & 91.18 & 92.63 & 90.98 \\
         
        \bottomrule
    \end{tabular}
    }
    \label{tab: deberta results}
\end{table*}

%% file: sec/6.conclusion.tex
\section{Conclusion}

This paper introduces a novel method for extending matrix-based parameter-efficient fine-tuning (PEFT) techniques to higher-dimensional parameter spaces, preserving their structural integrity. 
By treating weight updates as perturbations within the framework of Lie groups, we ensure that modifications respect the spatial and structural properties of high-dimensional tensors, such as convolutional layers.
Our experiments demonstrate the effectiveness of this approach, achieving superior performance in both computational efficiency and task-specific adaptation. 
This method expands the applicability of PEFT techniques to a broader range of architectures, offering a scalable solution for fine-tuning large foundation models while maintaining their structural properties.
Future work will explore further theoretical developments and extend this framework to a wider array of architectures, enhancing its scalability and robustness.

%% file: sec/supp.tex
\section{Details for Fine-tuning ConvNeXt-V2-B}

Table \ref{tab:training convnext} provides a set of custom hyperparameters used in all experiments.
For additional default parameters that are not included in Table \ref{tab:training convnext}, please refer to the MMDetection repositories.

\begin{table*}[ht]
    \centering
    \caption{Hyper-parameter setup for object detection and segmentation.}
    \resizebox{0.99\textwidth}{!}{
    \begin{tabular}{c c c c c  c c c}
    \toprule
    \textbf{Dataset} & \textbf{Toolkit} & \textbf{Model} & \textbf{Schedule} & \textbf{LR} & \textbf{BS} & \textbf{Optimizer} & \textbf{Weight Decay}   \\\hline
    COCO & MMDetection~\cite{chen2019mmdetection} & Mask R-CNN~\cite{mask_rcnn} & 12ep & 1e-4 & 32 & AdamW & 5e-2 \\
    \bottomrule
    \end{tabular}}
    \label{tab:training convnext}
\end{table*}

We mainly adopt the VTAB-1K Benchmark for image classification. 
VTAB-1K \cite{zhai2019visual} consists of 19 image classification tasks that cover a broad range of domains, divided into three categories: Natural, Specialized, and Structured. 
These tasks encompass a wide array of potential downstream applications, making the benchmark a robust indicator of a method’s transfer learning abilities.
Each dataset is composed of 800 training samples and 200 validation samples.
In line with previous work \cite{jia2022visual,jie2023fact}, we fine-tune the pre-trained ConvNeXt-V2-B model using all 1,000 training and validation samples and evaluate its performance on the test set.
Consistent with \cite{jia2022visual, lian2022scaling}, we use unnormalized inputs, as done in the original VTAB paper \cite{zhai2019visual}.

\section{Details for Generation Tasks}
We fine-tune text-to-image diffusion model, SDXL5 \cite{podell2023sdxl}, designed for subject-specific generation tasks, as presented in recent work \cite{ruiz2023dreambooth}.
The goal of this task is to generate images that closely align with prompts associated with a specific subject, defined by a few example images. 
This process begins by fine-tuning a text-to-image model using image-text pairs, where the text includes a unique identifier (e.g., ``A picture of a [V] cat'').
Afterward, the model generates images based on new prompts that include this identifier, with the aim of creating images that reflect the learned subject.
Fine-tuning is performed with a learning rate of 1e-4 and a batch size of 4.
The model is trained for 500 steps on a single 80GB A100 GPU, taking approximately 21 minutes to finish. During the generation phase, we run 50 inference steps per prompt to produce the final images, which takes about 30 seconds.
We primarily use the official DreamBooth dataset \cite{ruiz2023dreambooth} for the diffusion process.

\section{Details for Commonsense Reasoning Tasks}
The commonsense reasoning benchmarks comprise 8 different sub-tasks, each with a distinct dataset, including BoolQ \cite{clark2019boolq}, PIQA \cite{bisk2020piqa}, SIQA \cite{sap2019socialiqa}, HellaS. \cite{zellers2019hellaswag}, WinoG. \cite{sakaguchi2021winogrande}, ARC-e/ARC-c \cite{clark2018thinkarce}, and OBQA \cite{mihaylov2018canobqa}. In accordance with the protocol described by \cite{hu2023llm}, we combine the training datasets from all sub-tasks to form the Commonsense170K dataset, and perform evaluations on each sub-task’s respective testing set.

We incorporate results from ChatGPT’s implementation using the gpt-3.5-turbo API, specifically focusing on zero-shot Chain of Thought approaches \cite{wei2022chain}.
For fair comparison, the initial fine-tuning for models utilizing LieRA is performed under LoRA configurations with the learning rate being the only variable optimized for better performance. The hyper-parameter settings for LieRA are shown in Table \ref{tab: cr detail}. 
The results for LoRA are taken from \cite{hu2023llm,liu2024dora}.

\begin{table*}[!ht]

\centering
\renewcommand\arraystretch{1}
\setlength{\tabcolsep}{15mm}
\caption{Hyper-parameter settings on commonsense reasoning tasks.}
\resizebox{\textwidth}{!}{
\begin{tabular}{c | c c | c c} 

\toprule

\textbf{Hyper-parameters} & \multicolumn{2}{c}{LLaMA-7B} & \multicolumn{2}{c}{LLaMA3-8B}\\ 

\midrule

Rank $r$ & 16 & 32 & 16 & 32 \\ \midrule

$\alpha$ & 32 & 64 & 32 & 64 \\ \midrule

LR & 2e-3 & 1e-3 & 6e-4 & 8e-4 \\ \midrule

LR Scheduler & \multicolumn{4}{c}{Linear} \\ \midrule

Dropout & \multicolumn{4}{c}{0.05} \\ \midrule

Optimizer & \multicolumn{4}{c}{AdamW} \\ \midrule

Batch size & \multicolumn{4}{c}{16} \\ \midrule

Warmup Steps & \multicolumn{4}{c}{100} \\ \midrule

Epochs & \multicolumn{4}{c}{3} \\ \midrule

Where & \multicolumn{4}{c}{Q, K, V, Up, Down} \\

\bottomrule

\end{tabular}}
\label{tab: cr detail}
\end{table*}

\section{Details for NLU Tasks}

For the natural language understanding (NLU) task, we use the General Language Understanding Evaluation (GLUE) benchmark \cite{wang2018glue}, which is designed to assess performance across a variety of tasks.
The benchmark includes two single-sentence classification tasks, CoLA \cite{warstadt2019neural} and SST-2 \cite{socher2013recursive}, three similarity and paraphrase tasks: MRPC \cite{dolan2005automatically}, QQP \cite{wang2018glue}, and STS-B \cite{cer2017semeval}, and three natural language inference tasks: MNLI \cite{williams2017broad}, QNLI \cite{rajpurkar2016squad}, and RTE \cite{dagan2005pascal, bar2006second, giampiccolo2007third, bentivogli2009fifth}.
We fine-tune both the DeBERTaV3-base and DeBERTaV3-large models \cite{he2021debertav3} for this task. The corresponding hyper-parameter settings are listed in Table \ref{tab: nlu detail}.

\begin{table*}
\centering
\caption{Hyper-parameter settings on NLU task.}
\setlength{\tabcolsep}{4mm}
\resizebox{\textwidth}{!}{
\begin{tabular}{c | c c c c c c c c} 

\toprule

Hyper-parameter & MNLI & SST-2 & CoLA & QQP & QNLI & RTE & MRPC & STS-B\\ 

\midrule

Optimizer & \multicolumn{8}{c}{AdamW} \\ \midrule

Warmup Ratio & \multicolumn{8}{c}{0.1} \\ \midrule

LR schedule & \multicolumn{8}{c}{Linear} \\  \midrule

Rank $r$ & \multicolumn{8}{c}{2 \& 8}\\ \midrule

LoRA alpha & \multicolumn{8}{c}{4 \& 16} \\ \midrule

Max Seq. Len. & 256 & 128 & 64 & 320 & 512 & 320 & 320 & 128 \\ \midrule

Batch Size & 32 & 32 & 32 & 32 & 32 & 32 & 32 & 32 \\ \midrule

Learning Rate & 5e-3 & 8e-3 & 8e-3 & 1e-3 & 5e-3 & 2e-3 & 1e-3 & 5e-3 \\ \midrule

Epochs & 12 & 24 & 25 & 5 & 5 & 50 & 30 & 25  \\ 

\bottomrule

\end{tabular}}
\label{tab: nlu detail}
\end{table*}